\def\year{2018}\relax
\documentclass[letterpaper]{article} 
\usepackage{aaai18}  
\usepackage{times}  
\usepackage{helvet}  
\usepackage{courier}  
\usepackage{url}  
\usepackage{graphicx}  
\usepackage{xcolor}

\usepackage{latexsym}

\usepackage{amsmath}
\usepackage{amssymb}

\usepackage{subcaption}

\usepackage{enumitem}

\usepackage[ruled,vlined,linesnumbered]{algorithm2e}

\usepackage{natbib}

\usepackage[polish,english]{babel}
\usepackage[T1,nomathsymbols]{polski}
\usepackage[utf8]{inputenc}

\usepackage[pdftitle={TODO},
pdfauthor={Jakob Suchan, Mehul Bhatt},
linkbordercolor=white,
bookmarksopen=false,
bookmarksnumbered=false]{hyperref}

\hypersetup{
    pdftitle={Visual Explanation by High-Level Abduction: On Answer-Set Programming Driven Reasoning about Moving Objects}, 
    pdfauthor={Jakob Suchan, Mehul Bhatt, Przemek Walega, Carl Schultz},     	
    pdfsubject={artificial intelligence, knowledge representation and reasoning, computer vision, visual perception}, 
    pdfkeywords={TODO} {TODO} {TODO}, 					
    unicode=false,          									
    pdftoolbar=false,        									
    pdfmenubar=true,        								
    pdffitwindow=false,     								
    pdfstartview={FitH},    									
    pdfnewwindow=true,      								
    bookmarks=true,         								
    colorlinks=true,       									
    linkcolor=red!70!black,          							
    citecolor=blue!90!black,        							
    filecolor=blue,      									
    urlcolor=blue!90!black          								
}

\newcommand{\Pred}[1]{{\mathsf{\lowercase{#1}}}}

\newcommand{\aspLong}[0]{Answer Set Programming}
\newcommand{\aspShort}[0]{ASP}

\newcommand{\VXP}[1]{{{$\mathcal{VXP}_{#1}$}}}

\newcommand{\abd}[0]{$\Sigma_{abd}$}
\newcommand{\track}[0]{$\Sigma_{trk}$}

\newcommand{\predThF}[1]{{\operatorname{\mathsf{#1}}}}

\newcommand{\objectSort}[0]{$\mathcal{\color{blue!96!black}O}$}

\newcommand{\secTitle}[1]{{{#1}}}

\definecolor{YellowGreen}{RGB}{160,200,40}
\definecolor{mathcolor}{RGB}{7,72,110}

\newcommand{\bulletpoint}[1]{\null\quad -- $~$ \begin{minipage}[t]{0.9\columnwidth}{#1}\end{minipage}\\[2pt]}
\newcommand{\bulletpointb}[1]{\null\quad -- $~$ \begin{minipage}[t]{0.9\columnwidth}{#1}\end{minipage}}

\begin{document}

%

%
%
%
%
%

\title{{\sffamily{Visual Explanation by High-Level Abduction}}\\[3pt]{\textnormal{\large\sffamily\upshape{On Answer-Set Programming Driven Reasoning about Moving Objects}}}}

\author{Jakob Suchan$^1$, Mehul Bhatt$^{1,2}$, Przemys{\l}aw Wa{\l}\k{e}ga$^3$, \and Carl Schultz$^4$\\[4pt]
Cognitive Vision -- \url{www.cognitive-vision.org}\\[2pt]EASE CRC -- \url{http://ease-crc.org}\\[4pt]$^1$HCC Lab., University of Bremen, Germany, $^2$MPI Lab., \"{O}rebro University, Sweden\\
$^3$University of Warsaw, Poland, and $^4$Aarhus University, Denmark
}

\maketitle

\begin{abstract}
We propose a hybrid architecture for systematically computing robust visual explanation(s) encompassing hypothesis formation, belief revision, and default reasoning with video data. The architecture consists of two tightly integrated synergistic components: {\bf\small(1)} (functional) answer set programming based abductive reasoning with {\scriptsize\sffamily \textsc{space-time tracklets}} as native entities;  and {\bf\small(2)}  a visual processing pipeline for detection based object tracking and motion analysis. 

We present the formal framework, its general implementation as a (declarative) method in answer set programming, and an example application and evaluation based on two diverse video datasets: the MOTChallenge benchmark developed by the vision community, and a recently developed Movie Dataset.
\end{abstract}


\section{\secTitle{Introduction}}
A range of empirical research areas such as cognitive psychology and visual perception articulate human visual sensemaking as an inherently abductive (reasoning) process \citep{Moriarty96,Magnani2015} involving tight linkages between low-level sub-symbolic processes on the one hand, and high-level object and event-based segmentation and inference involving concepts and relations on the other. In spite of the state of the art in artificial intelligence and computer vision, and most recent advances in neural visual processing, generalised \emph{explainable visual perception} with conceptual categories in the context of dynamic visuo-spatial imagery remains an exceptionally challenging problem presenting 
many research opportunities at the interface of {Logic, Language, and Computer Vision}.

\subsubsection{Explainable Visual Perception} 
We define explainable visual perception from a human-centred, and commonsense reasoning viewpoint. In this paper, it denotes the ability to declaratively:

\smallskip

\noindent  {\color{blue}\small\VXP{1}}:\quad  hypothesise spatio-temporal belief (states) and events; events may be both primitive or temporally-ordered aggregates; from a more foundational viewpoint, what is alluded to here is a robust mechanism for  \emph{counterfactual} reasoning.

\smallskip

\noindent  {\color{blue}\small\VXP{2}}:\quad   revise spatio-temporal beliefs, e.g., by non-monotonically updating conflicting knowledge, to fix inherently incompatible configurations in {space-time} defying geometric constraints and commonsense laws of naive physics, e.g., pertaining to physical (un)realisability, spatio-temporal continuity.

\smallskip

\noindent  {\color{blue}\small\VXP{3}}:\quad   make default assumptions, e.g., about spatio-temporal property persistence concerning occupancy or position of objects; identity of tracked objects  in space-time. 

\smallskip


\noindent Explanatory reasoning in general is one of the hallmarks of general human reasoning ability; robust explainable visual perception particularly stands out as a foundational functional capability within the human visuo-spatial perception faculty. In this respect, the following considerations ---establishing the scope of this paper--- are important wrt. {\color{blue}\small\VXP{1-3}} :

\begin{itemize}

	\item our notion of explainability is driven by the ability to support commonsense, semantic question-answering over dynamic visuo-spatial imagery within a declarative KR setting; 

	\item the features alluded to in {\small\VXP{1-3}} are not exhaustive; we focus on those aspects that we deem most essential for the particular case of \emph{movement tracking}.
	
\end{itemize}	

\begin{figure*}
\center
\includegraphics[width = 0.9\textwidth]{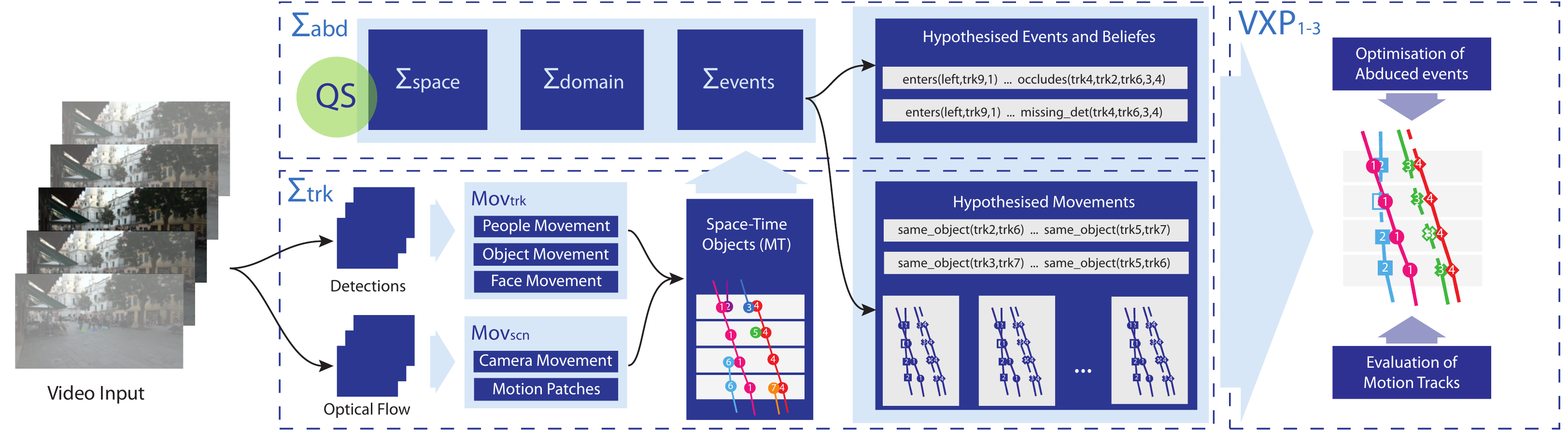}
\caption{\small\sffamily \textbf{Visual Explanation} -- A Hybrid Architecture Integrating Low-Level Visual Tracking and High-Level Abduction}
\label{fig:visual_abduction_pipeline}
\end{figure*}

\subsubsection{A Hybrid Architecture for Visual Explanation} 
This paper is driven by the development of a \emph{visual explanation component} within a large-scale computational  vision \& perception system targeted at a range of cognitive interaction technologies and autonomous systems where dynamic visuo-spatial imagery is inherent. 


The key contribution is a hybrid visual explanation method based on the integration of high-level abductive reasoning within \aspLong\ (\aspShort) (\cite{Brewka:2011:ASP}) on the one hand, and low-level visual processing for object tracking on the other. The core focus of the paper is on the theory, implementation, and applied evaluation of the  visual explanation method. We particularly emphasise the closely-knit nature of two key sub-components representing abductive explanation (\abd) and low-level motion tracking (\track) modules respectively:

\smallskip

\noindent{\color{blue}\abd}.\quad  \aspShort-based abductive reasoning with abstract visuo-spatial concepts ---such as  {\scriptsize\sffamily \textsc{objects, events, space-time tracklets}}--- as native objects within \aspShort

\smallskip

\noindent{\color{blue}\track}.\quad   Low-level visual processing pipeline for motion tracking, consisting of detection-based object tracking and optical-flow based (scene-level) movement tracking

\smallskip

\noindent The abductive component \abd\ is suited for a wide-range of dynamic visuo-spatial imagery; however, we only focus on {\sffamily video} in this paper. As an application, we focus on scene interpretation from video with two datasets:  a \emph{Movie Dataset} \citep{MovieDataset-Geometry} and the \emph{MOT16 Dataset}, a benchmark dataset released as part of The Multiple Object Tracking Challenge \citep{MOT16-Benchmark}.

\section{\secTitle{Visual Explanation: A Hybrid Architecture}}
We present a general theory for explaining visuo-spatial observations by integrating low-level visual processing and high-level abductive reasoning (Fig. \ref{fig:visual_abduction_pipeline}). As such, we consider visual abduction as reasoning from visual observations to explanations consisting high-level events grounded in low-level motion tracks. 
The resulting set of hypotheses is optimised based on the abduced events and the corresponding object movement.


\subsection{Ontology: Space, Time, Objects, Events}
The framework for abducing visual explanations is based on visuo-spatial domain objects 
representing the visual elements in the scene. 
The domain objects are associated with spatio-temporal objects describing motion tracks obtained from \track, which form the basis for qualitative spatio-temporal abstractions facilitating high-level reasoning about visuo-spatial dynamics.

\smallskip

\noindent The \textbf{Qualitative Spatio-Temporal Domain} ({\color{blue}$\mathcal{QS}$}) is characterised by the basic spatial and temporal  entities ({\color{blue}$\mathcal{E}$}) that can be used as abstract representations of domain-objects and the relational spatio-temporal structure ({\color{blue}$\mathcal{R}$}) that characterises the qualitative spatio-temporal relationships amongst the supported entities in ({\color{black}$\mathcal{E}$}). 
For this paper, we restrict the basic spatial entities to:\\[4pt]
\bulletpoint{\emph{points} are a pair of reals $x,y$, }
\bulletpoint{\emph{axis-alined rectangles} are a point $p$ and its width and height $w,h$,}

and the temporal entities to:\\[4pt]
\bulletpointb{\emph{time-points} are a real $t$}

\noindent Visuo-spatial \textbf{domain objects} \objectSort\ = $\{o_1, o_2, ... , o_n\}$ are described as spatio-temporal objects by a set of spatial entities, i.e.,  \emph{points}, and \emph{axis-aligned rectangles}, in time. 
Towards this, $\color{blue}\mathcal{MT}$ contains all object tracks obtained form \track.
The track of a single object $o_i$ is represented by $\mathcal{MT}_{o_i}$ = ($\varepsilon_{t_{s}}, ..., \varepsilon_{t_{e}}$), where $t_s$ and $t_e $ denote the start and end frame of the track and $\varepsilon_{t_s}$ to $\varepsilon_{t_e}$ denotes a spatial primitive representing the object $o_i$ at the time points $t_s$ to $t_e $, e.g., the axis aligned bounding box of the object.

\smallskip

For reasoning about visuo-spatial phenomena of object tracks, spatio-temporal relationships ($\mathcal{R}$) between the basic entities in $\mathcal{E}$ may be characterised with respect to arbitrary spatial and spatio-temporal domains such as \emph{mereotopology, orientation, distance, size, motion}. 
From the viewpoint of the examples of this paper, it suffices to focus on the language of the mereotopological system of the Region Connection Calculus (RCC8) \citep{randell92RCC-small} consisting of the following jointly exhaustive and pair-wise disjoint relations: {disconnected} ($\Pred{DC}$), {externally connnected} ($\Pred{EC}$), {partially overlapping} ($\Pred{PO}$), {equal} ($\Pred{EQ}$), {(non\nobreakdash-) tangential proper part} ($\Pred{(N)TPP}$), and their inverse  ($\Pred{(N)TPPi}$). 

%

\smallskip

Abducable \textbf{events} ($\Theta$) and \textbf{beliefs} ($\Phi$) are defined by their (spatio-temporal) \emph{preconditions} and \emph{observer effects}, i.e., for each event $\theta \in \Theta$ we define which properties of the scene have to be true for the event to be possible, and what the (visible) effects of the event are. 
In the case of visual abduction, properties of the scene are determined by the visually observed object tracks and represent qualitative relations between tracks, i.e., spatial relation $r \in  \mathcal{R}$ holding between basic spatial entities  $\varepsilon$ of a motion track.
Complex events are defined by combining multiple events and beliefs, e.g., an event of an object $o_i$ passing behind another object $o_j$ can be defined based on the events of $o_i$ being occluded by $o_j$ and $o_i$ and $o_j$ changing sides.

\subsection{Abducing Visual Explanations}

We implement the theory for visual explanations combining visual processing for object detection and tracking, and estimating movements in the scene, with ASP based reasoning about events, objects, and spatial-dynamics (Fig. \ref{fig:visual_abduction_pipeline}).
The main components of the overall tightly-integrated system comprising of low-level motion tracking with high-level explanation  is as follows:

\smallskip

\noindent \textbf{I}. \emph{Visuo-Spatial Observations} ({\color{blue}$\mathcal{VO}$}) $~$ -- $~$ low-level visual processing
consisting of \emph{detection based tracking} of object and people movements. 

\smallskip

\noindent \textbf{II}. \emph{Hypotheses} ($\color{blue}\mathcal{H}$) $~$ -- $~$ abducing hypotheses including belief states, events, and default assumptions given a set of visuo-spatial observations ({\color{black}$\mathcal{VO}$}). 

\smallskip

\noindent \textbf{III}. \emph{Hypotheses to Explanations} --- as encompassed in {\small\VXP{1-3}}--- are generated by evaluating abduced hypothesis ($\mathcal{H}$) based on high-level optimisation of event sequences and low-level cost minimisation of corresponding motion tracks. 

\smallskip

\subsubsection{\secTitle{I.\quad Visuo-Spatial Observations} ({\color{blue}$\mathcal{VO}$})} 
Visual explanations are based on observations obtained from visuo-spatial imagery, e.g., video, RGB-D.
For the examples in this paper, we focus on \emph{detection and tracking} of people and objects for estimating motion trajectories of semantic entities in the scene. 
However, the presented approach is also capable of incorporating other kinds of motion, e.g., optical flow based \emph{low-level movement analysis} using long term observations \citep{DBLP:journals/pami/OchsMB14}, or dense motion-tracklets \citep{DBLP:journals/ijcv/GaidonHS14} for estimating pixel level motion, corresponding to \emph{camera movement}, or \emph{fine grained object motion}, etc. 
This may be used to abduce fine grained interactions and the interplay of different movements, e.g. people movement in the presence of camera movement, by combining motion trajectories of semantic entities with pixel movements.

\smallskip

\noindent \textit{Movement}  of people and objects is estimated following the \emph{tracking by detection} paradigm, which is based on object detections for each frame and association of the detections across frames. 
Object detections can in principle be obtained using any state of the art (deep learning based) detector (e.g., \emph{faster RCNN} \citep{DBLP:conf/nips/RenHGS15}, YOLO \citep{DBLP:conf/cvpr/RedmonDGF16}), or deformable part models (DPM) \citep{Felzenszwalb2010}. For the examples in this paper we are using \emph{faster RCNN} in the movie examples and DPM detections for the MOT dataset (which come as part of the dataset).
For association of detections we apply the well established approach of combining \emph{min cost assignment} and \emph{kalman filters} for finding optimal tracklets, where the cost for assigning a detection to a track is calculated by the distance between the prediction for a track and the detection. 

\begin{itemize}

\item \emph{Prediction} \quad for each track  \emph{Kalman filters} are used to predict the next position of the track and the costs for each detection is calculated based on the distance between the prediction and the detection.
 
\item \emph{Assignment} \quad   detections are assigned to a track using \emph{min cost assignment} which calculates the best assignment of detection to tracks based on the costs calculated in the prediction step. If no assignment is possible for a detection a new track is started.

\end{itemize}

\noindent The resulting object tracks $\mathcal{MT}$ form the basis for abducing explanations on movement events occurring in the input data.

\smallskip

\subsubsection{\secTitle{II.\quad Hypotheses} ($\color{blue}\mathcal{H}$)}
Explanations for visual observations are abduced based on a sequence of visual observations obtained from the video data. For abducing visual explanations from $ \mathcal{VO}$, given,\\[4pt]
\bulletpoint{set  $ \mathcal{VO}$  consisting of visuo-spatial observations obtained from \track, }
\bulletpoint{domain independent theory of space and time ($\Sigma_{space}$) based on the spatio-temporal ontology ($\mathcal{QS}$)}
\bulletpoint{observable events ($\Sigma_{events}$) }
\bulletpoint{domain dependent background knowledge, describing properties of the domain ($\Sigma_{domain}$ )}

\noindent the task of visual abduction is to find a set of logically consistent hypotheses $\mathcal{H}$ consisting of high-level events and beliefs grounded in low-level motion tracks, such that:

\smallskip

\begin{center}
{{$ \Sigma_{space} \wedge \Sigma_{events} \wedge \Sigma_{domain} \wedge \mathcal{H} \models \mathcal{VO} $}}
\end{center}

\smallskip

\noindent The computed hypotheses ($\mathcal{H}$) are based on abducibles constituting primitive events and beliefs:  $\mathcal{H} \equiv \mathcal{H}_{Events} \wedge \mathcal{H}_{Belief}$; these hypotheses in turn are directly usable for inducing motion tracks: 

\begin{center}
$\mathcal{MT}_{\mathcal{VXP}} \longleftarrow \mathcal{H}_{{event}} \wedge\mathcal{H}_{{belief}}\wedge\mathcal{MT}$
\end{center}

\noindent The resulting motion tracks $\mathcal{MT}_{\mathcal{VXP}}$ represent the low-level instantiation of the abduced high-level event sequence.

\subsubsection{\secTitle{III.\quad Hypotheses to Explanations}}

Hypotheses for visual observations ($\mathcal{VO}$) may be ranked based on the abduced event sequences and cost minimisation of corresponding motion trajectories, i.e., the costs for connecting motion tracks in the hypothesised movements, e.g. considering changes in \emph{velocity}, \emph{size}, and \emph{length} of missing detections. 
As such, hypothesised explanations are ranked using the built in optimisation functionality of ASP\footnote{For optimisation we use ASP with the so-called \emph{weak constraints} (\cite{gekakasc12a}), i.e., constraints  whose violation has a predefined cost. When solving an ASP program with weak constraints, a search for an answer set with a minimal cost of violated constraints is performed. Each such minimal-cost answer set is called optimal. The mechanism involving weak constraints enables us to set preferences among hypothesised explanations and search for the ones that are most preferred (optimal). Importantly, the approach enables us to exhaustively search for all optimal explanations. As a result, we can subsequently use other (more fine-graded) evaluation techniques to choose the most preferred explanations.}. In particular, we use minimisation by assigning preferences to the abducables events and beliefs and optimise towards minimising the costs of events and beliefs in the answer. E.g., by minimising the duration of missing detections for a particular object, or minimising assigning the property noise to a track to explain its observation.

\begin{itemize}

\item \emph{High-level event sequences} \quad the cost for high-level events is estimated by assigning a cost for each event. Additionally, for events having a duration there is also a cost assigned to the length of the event, e.g., to abduce that a track is noise is more likely, when it is a very short track. 
These costs are weighted based on the abduced event that caused the missing detections, e.g., missing detections caused by an occlusion are more likely to be longer (and therefore have a lower cost), than missing detections caused by the detector.

\item \emph{Low-level motion characteristics} \quad the cost of the motion tracks $\mathcal{MT}$ is estimated based on the characteristics of the abduced movement. Towards this we consider changes in \emph{velocity}, for each abduced event that connects two object tracks. For the examples in this paper we use a constant velocity model to minimize changes in velocity of an abduced object track.

\end{itemize}

\noindent The best explanation is selected by \emph{minimising} the costs of the hypothesised answer set based on the motion and the high-level event sequence.  
The final movement tracks for the optimal explanation $\mathcal{MT}_{\mathcal{VXP}}$ are then generated by predicting the motion of the object for each hypothesised event associating two tracks, using linear interpolation.

\begin{table}
\begin{center}
\scriptsize
\begin{tabular}{|l|p{1.45in}|}
\hline
\textbf{EVENTS} & \textbf{Description} \\\hline

$\predThF{enters(Border, Trk, T)}$ & The object corresponding to track $\mathsf{Trk}$ enters the scene at time point $\mathsf{T}$.\\[2pt]

$\predThF{exits(Border, Trk, T)}$ & The object corresponding to track  $\mathsf{Trk}$ exits the scene at time point $\mathsf{T}$.\\[2pt]

$\predThF{occludes(Trk_1, Trk_2, Trk_3, T_1, T_2)}$ & The object corresponding to track $\mathsf{Trk_1}$ and track $\mathsf{Trk_2}$ is occluded by the object corresponding to track $\mathsf{Trk_3}$ between time points $\mathsf{T_1}$ and $\mathsf{T_2}$.\\[2pt]

$\predThF{missing\_det(Trk_1, Trk_2, T_1, T_2)}$ & Missing detections for the object corresponding to the tracks $\mathsf{Trk_1}$ and $\mathsf{Trk_2}$ between time points $\mathsf{T_1}$ and $\mathsf{T_2}$.\\\hline

\hline
\textbf{COMPLEX EVENTS} & \textbf{Description} \\\hline

$\predThF{passing\_behind(O_1, O_2, T_1, T_2)}$ & Object $\mathsf{O_1}$ is passing behind object $\mathsf{O_2}$ between time points $\mathsf{T_1}$ and $\mathsf{T_2}$.\\[2pt]

$\predThF{moving\_together(O_1, O_2, T_1, T_2)}$ &  Objects $\mathsf{O_1}$ and $\mathsf{O_2}$  are moving together between time points $\mathsf{T_1}$ and $\mathsf{T_2}$.\\
\hline

\hline
\textbf{BELIEFS} & \textbf{Description} \\\hline

$\predThF{same\_object(Trk_1, Trk_2)}$ & The tracks  $\mathsf{Trk_1}$ and  $\mathsf{Trk_2}$ belong to the same object.\\[2pt]

$\predThF{belongs\_to(Trk_1, Trk_2)}$ & The object corresponding to track $\mathsf{Trk_1}$ is a part of the object corresponding to track $\mathsf{Trk_2}$.\\[2pt]

$\predThF{noise(Trk)}$ & Track $\mathsf{Trk}$ is a faulty detection.\\\hline

\end{tabular}
\caption{\small\sffamily \textbf{Abducibles}: Events and Beliefs for Explaining Observed Object Tracks.}
\label{tbl:events}
\end{center}
\end{table}%

\section{Visuo-Spatial Phenomena}\label{sec:visuo-spatial_phenomena}

The framework may be used for abducing explanations by modelling visuo-spatial phenomena including but not limited to:

\begin{itemize}

\item\emph{Object Persistence} \quad objects can not appear and disappear without a cause, e.g. getting occluded, leaving the field of view of the camera, etc.

\item\emph{Occlusion} \quad objects may disappear or re-appear as a result of occlusion between two non-opaque objects. 

\item\emph{Linkage} \quad objects linked to each other, such that movement of one object influences movement of the other object, e.g. a face belonging to a person.

\item\emph{Sensor Noise} \quad observations that are based on faulty data, e.g. missing information, miss-detections, etc.
	
\end{itemize}

\smallskip

\begin{figure*}[t]
\centering
\includegraphics[width = 0.9\textwidth]{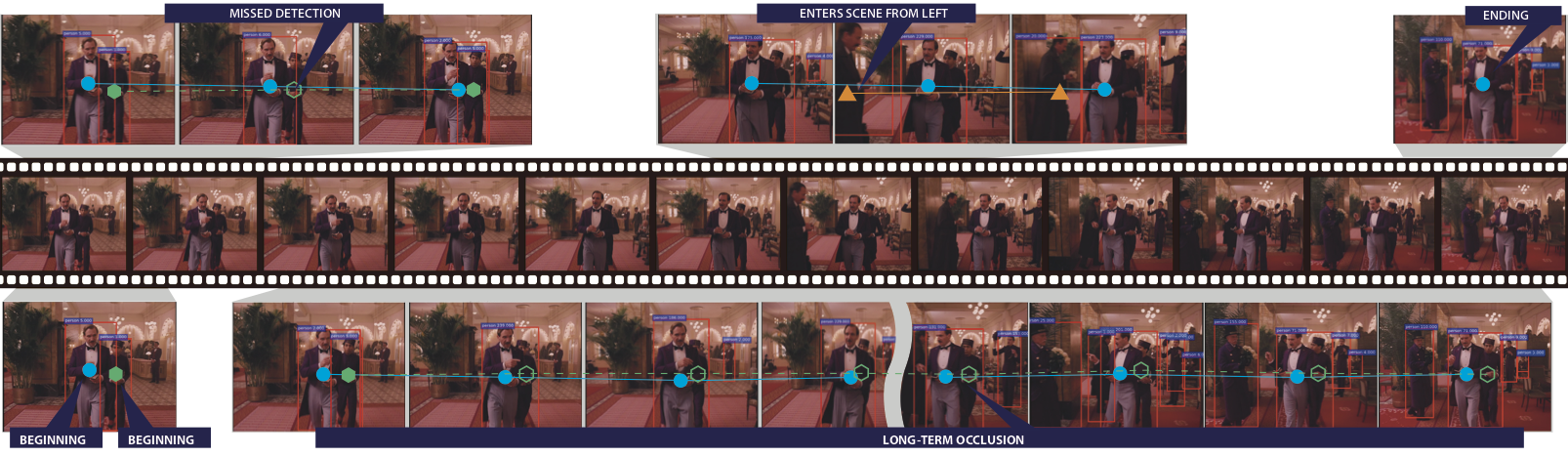}
\caption{\small\sffamily \textbf{People Movement} -- Scene from the movie The Grand Budapest Hotel (2014) by Wes Anderson}
\label{fig:qualitative_results}
\end{figure*}

\subsubsection{\secTitle{Event Semantics as Spatial Constraints}}
For explaining perceived visuo-spatial dynamics of objects in the scene, we define the basic events listed in Table \ref{tbl:events} to assure spatio-temporal consistency, e.g. object persistence, or occlusion. The focus is on explaining appearance and disappearance of objects in the scene.
\footnote{The semantics of the underlying spatial and temporal relations with ({\color{blue}$\mathcal{QS}$})  is founded on the geometric and spatial reasoning capability provided by the ASPMT(QS) spatial reasoning system \citep{aspmtqs-lpnmr-2015}; the system, implemented within ASPMT \citep{joohyung-aspmt-2013}, is directly available to be used as a black-box within our visual explanation framework.}

\smallskip


\noindent $\blacktriangleright$\quad  \textit{Entering and Leaving} \quad Objects can only enter or exit the scene by leaving the screen at one of its borders.
For these events to happen the object has to be overlapping with the border of the screen while appearing or disappearing.

\smallskip

\noindent {\footnotesize\sffamily\color{blue!70!black} enters}:

\smallskip

%

\noindent \includegraphics[width = \columnwidth]{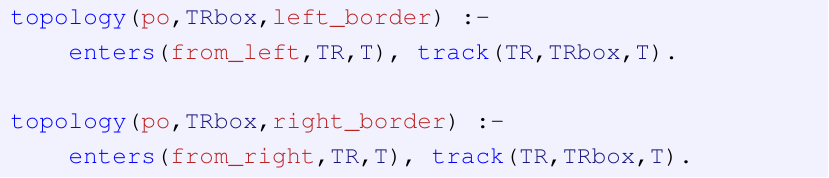}

\smallskip

\noindent {\footnotesize\sffamily\color{blue!70!black} exits}:

\smallskip

%

\noindent \includegraphics[width = \columnwidth]{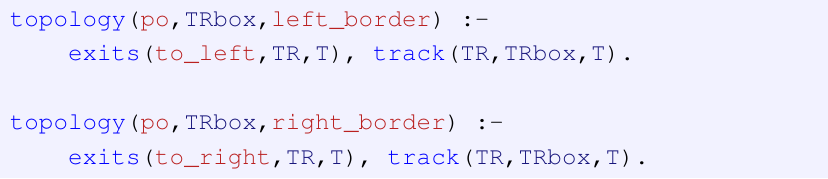}

\smallskip

\noindent  $\blacktriangleright$\quad  \textit{Missing Detections and Occlusion} \quad 
Appearance and disappearance of tracks in the middle of the screen can be either caused by a missing detection or by an occlusion from some other object.
The event that an object gets occluded by some other object may be possible, when the object disappears while overlapping with the other object.

\smallskip

\noindent {\footnotesize\sffamily\color{blue!70!black} occludes}:

\smallskip

%

\noindent \includegraphics[width = \columnwidth]{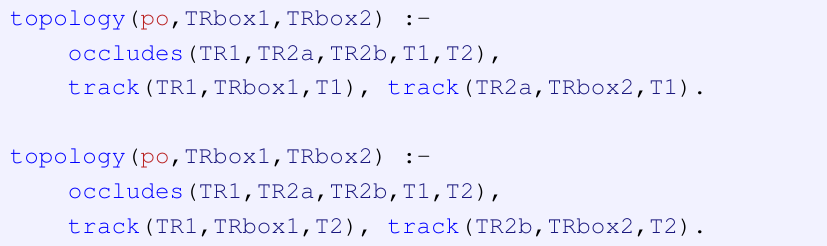}

\subsubsection{\secTitle{Generating Hypotheses on Events}}

We generate hypotheses explaining the observation of a track starting and ending based on the defined events, such that the spatial constraints defined above are satisfied. 

\smallskip

\begin{samepage}
\noindent {\footnotesize\sffamily\color{blue!70!black} starts}:
\nopagebreak

\smallskip


\noindent \includegraphics[width = \columnwidth]{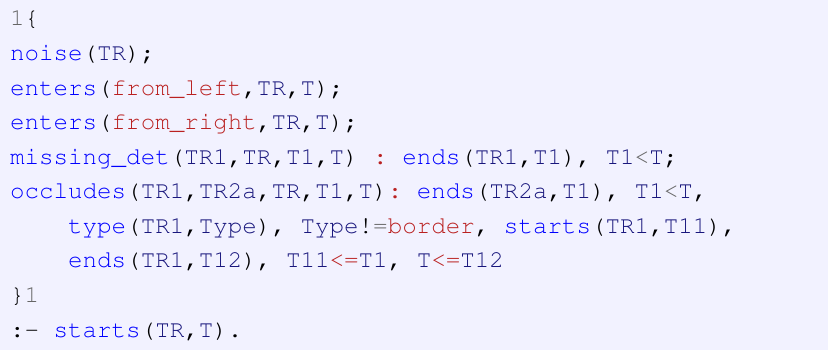}
\end{samepage}

\smallskip

\noindent {\footnotesize\sffamily\color{blue!70!black} ends}:

\smallskip


\noindent \includegraphics[width = \columnwidth]{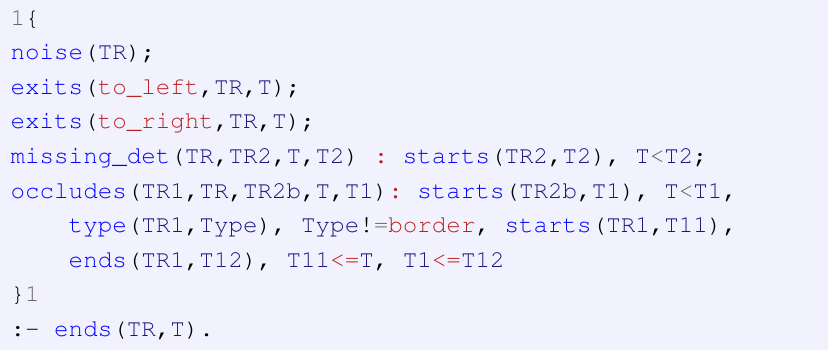}

\subsubsection{\secTitle{Beliefs as (Spatial) Constraints}} 

Beliefs about objects in the scene are stated as constraints in ASP. 

\smallskip

\noindent  $\blacktriangleright$\quad  \textit{Part-Whole Relations} \quad  E.g. the fact that every face belongs to exactly one person is stated as follows.

\smallskip


\noindent \includegraphics[width = \columnwidth]{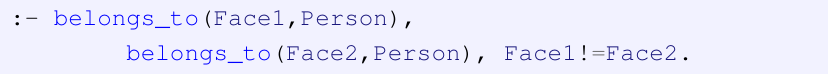}

\noindent Further we define that the face of a person has to stay together with the person it belongs to, using spatial constraints, i.e. the face track is a non-tangential proper part of the person track.

\smallskip


\noindent \includegraphics[width = \columnwidth]{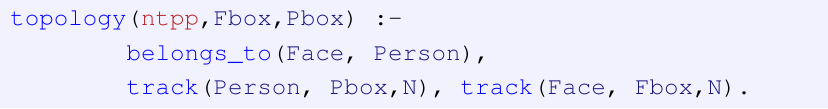}

\subsubsection{\secTitle{Generating Hypotheses on Beliefs}}

Hypotheses on faces belonging to persons are generated by stating that for each detected face, there has to be a corresponding person, such that the spatial constraint is satisfied.

\smallskip


\noindent \includegraphics[width = \columnwidth]{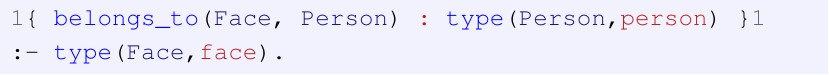}

\subsubsection{\secTitle{Costs of Hypotheses using Optimization}}

Costs for abduced visual explanations are minimized using ASP based optimization, e.g., the cost for missing detections are based on their length.

\smallskip


\noindent \includegraphics[width = \columnwidth]{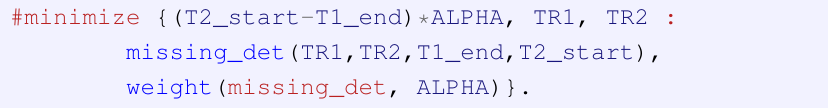}

\noindent Further, the characteristics of the underlying motion is taken into account, assuming constant velocity, by taking differences in velocity between the two tracks and the interpolated segment in between.

\smallskip


\noindent \includegraphics[width = \columnwidth]{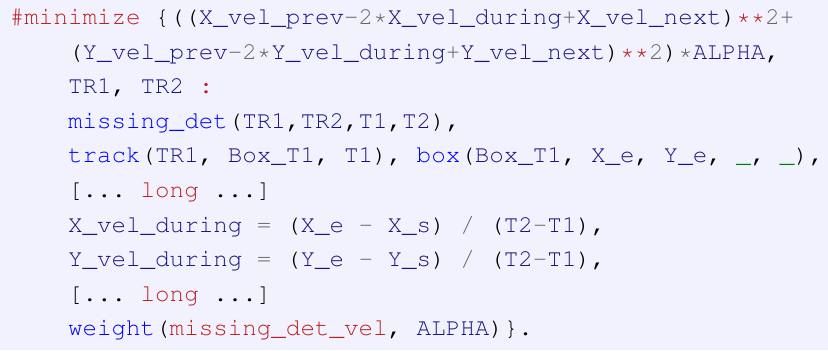}

\section{\secTitle{Application and Evaluation:\\Scene Interpretation with Moving Objects}}\label{sec:scene-interpret}

We demonstrate the proposed theory of visual abduction by applying it in the context of scene interpretation focussing on generating visual explanations on perceived motion. In particular, the emphasis is on spatio-temporal consistency of abduced explanations with respect to the underlying motion tracks. 

\medskip	

\noindent \textbf{Movie Dataset} \citep{MovieDataset-Geometry,filmDataset:2016:IJCAI}.\quad We use the video part of the Movie Dataset consisting of $16$ select scenes from $12$ films, with each scene ranging between $0:38$ minute to max. of $9:44$ minutes in duration. Most of the scenes involve multiple moving objects, and moving camera(s). Object detection with the movie dataset is performed using faster RCNN \citep{DBLP:conf/nips/RenHGS15} with the pre-trained VGG16 model for detection of people and objects in the scene.

\medskip

\noindent \textbf{Visual Explanation of Object Movement} \quad 
As an example consider the scene from the movie The Grand Budapest Hotel (2014) by Wes Anderson (Figure \ref{fig:qualitative_results}).  Here we abduce the movement of the two main characters walking down the hallway of the hotel. 
The set of visual observations consist of 11 tracks for the detected people in the scene. 
The abduced events explain occuring missing detections, occlusion and re-appearance, as well as entering, and leaving the scene.

\smallskip


\noindent \includegraphics[width = \columnwidth]{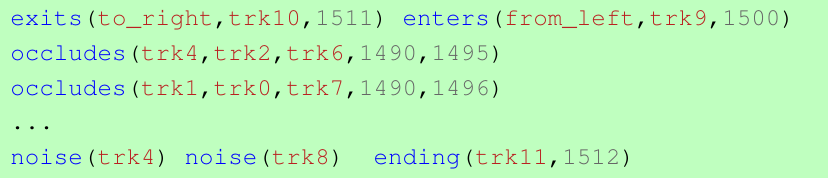}

\smallskip

\begin{figure}
\centering
\includegraphics[width = 1\columnwidth]{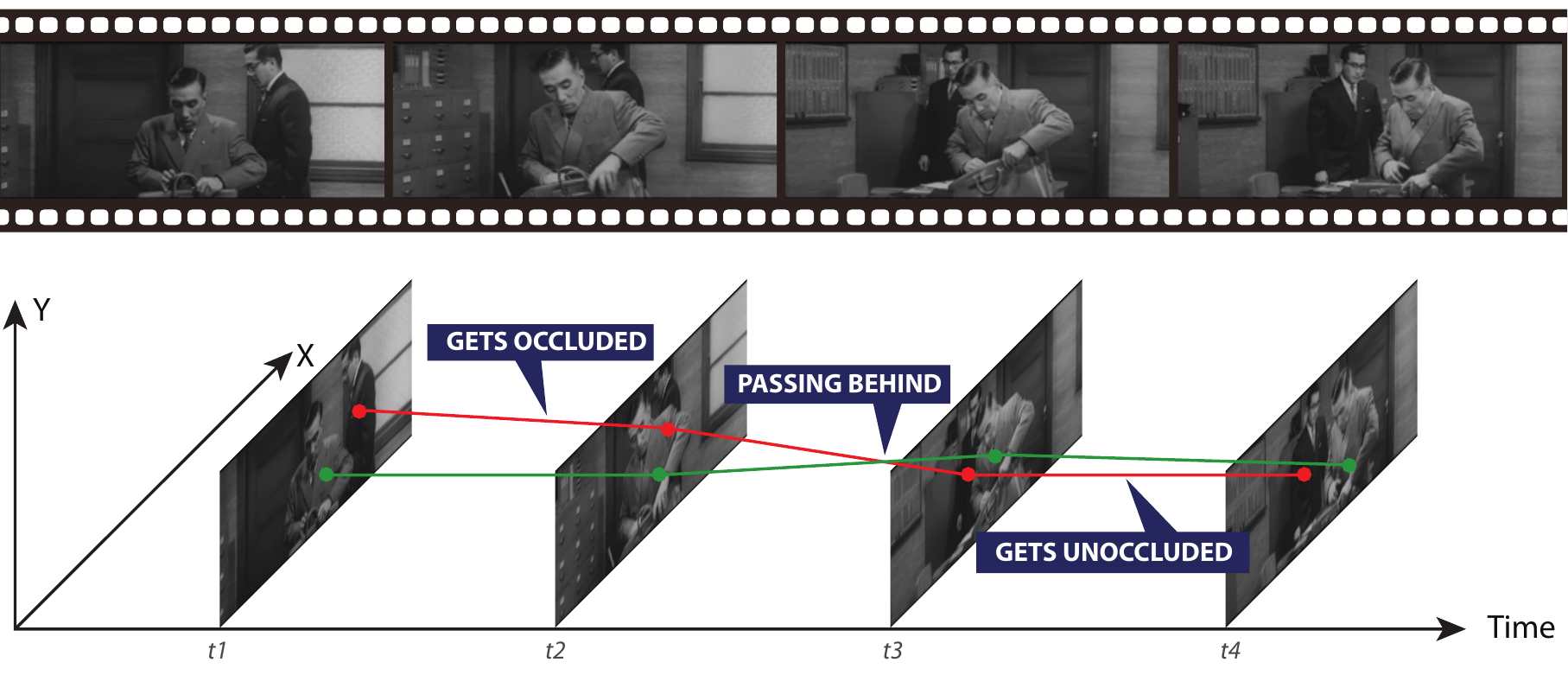}
\caption{\small\sffamily \textbf{Occlusion, while passing behind} -- Scene from ``The Bad Sleep Well'' by Akira Kurosawa (1960)}
\label{fig:BSW-example}
\end{figure}

\noindent Similarly we can abduce complex events based on movement events and belifes, e.g., based on the following sequence of movement events from the movie ``The Bad Sleep Well'' (1960) by Akira Kurosawa (depicted in Figure \ref{fig:BSW-example}), we can abduce the occurrence of the complex event \emph{passing behind} between two objects in the scene.
 
%

\noindent \includegraphics[width = \columnwidth]{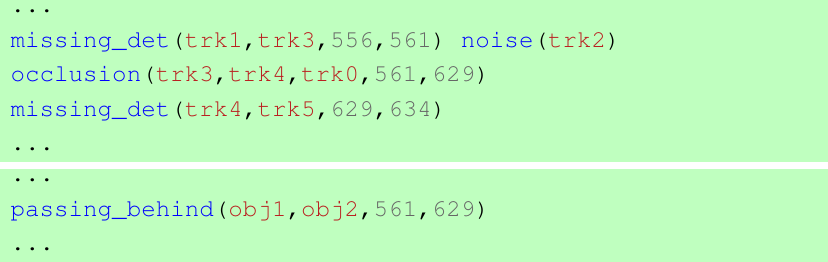}

\medskip

\noindent \textbf{Hypotheses on People and their Faces}\quad 
As an example for abducing properties of objects, we use face detections and abduce which face belongs to which person, using the part-whole relations defined in Section \emph{Visuo-Spatial Phenomena}.
%
%
The spatial constraint that a face track has to be inside a person track, is used to improve abduced object tracks.

\smallskip


\noindent \includegraphics[width = \columnwidth]{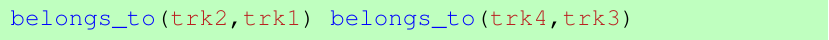}

\medskip

\begin{figure}
\centering
\includegraphics[width = 1\columnwidth]{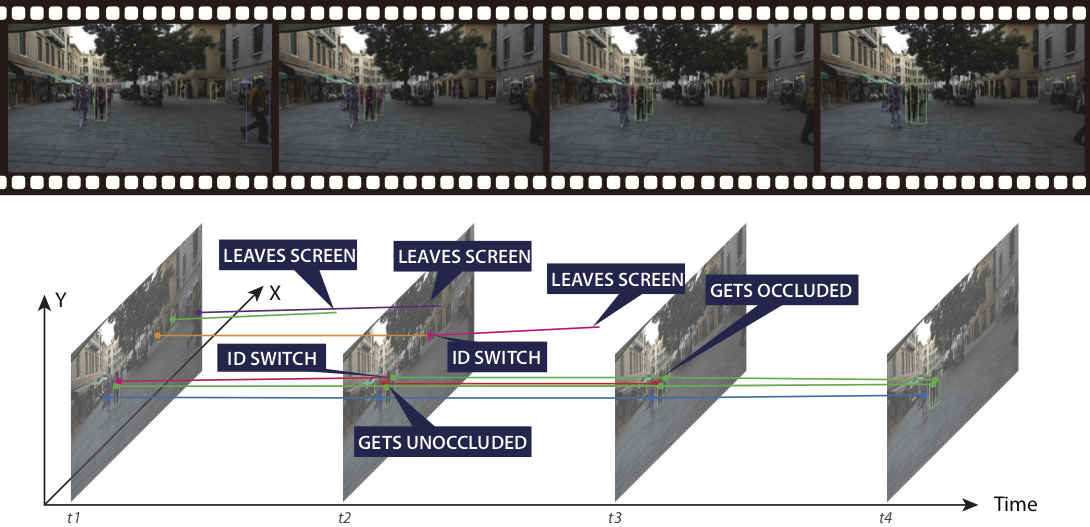}
\caption{\small\sffamily \textbf{Detection Errors} -- Scene from the MOT2016 dataset}
\label{fig:MOT-example}
\end{figure}

\noindent \textbf{\textbf{MOT16} Benchmark Video Dataset}\label{sec:dataset-eval}\quad We use the MOT16 \citep{MOT16-Benchmark} dataset consisting of highly accurate and consistent annotation protocols. MOT16 is a benchmark dataset released as part of The Multiple Object Tracking Challenge ({\small\sffamily MOTChallenge}). It consists of $14$ complex video sequences in highly unconstrained environments filmed with both static and moving cameras. We use the detections (provided by the dataset) based on deformable part models (DPM): these are noisy and include numerous miss detections, i.e. false positives and false negatives.  We focus on abducing people motion and on generating concise explanations for the perceived movements, i.e. under consideration of occlusion and appearance / disappearance of characters as per the abducible events in Table \ref{tbl:events}. As a result of the noisy detections and the complexity of the movements in the dataset the obtained motion tracks include a high amount of errors, e.g. identity switches, missing detections, etc. (Figure \ref{fig:MOT-example}). 
For the sample scene we abduced the following events:

\smallskip

%

\noindent \includegraphics[width = \columnwidth]{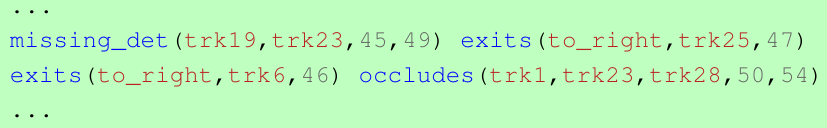}

\begin{table*}[t]
\begin{center}
\scriptsize

\begin{tabular}{|p{2.6cm}|p{1.8cm}|c|c|c|c|c|c|c|c|c|c|}
\hline
Sequence & Tracking & \textbf{MOTA} & \textbf{MOTP} & FP & M  & MM   & non-r. MM & r. MM  & TP & TR \\\hline 
\hline
\textbf{\sffamily \scriptsize The Bad Sleep Well}&
\scriptsize{ \textit{without} $\mathcal{VXP}$}&
58.5 \% &
80.8 \% &
1 &
86 &
5 &
0 &
5 &
0.875 &
1.0 \\

\tiny{(107 frames, 2 targets)} &
\scriptsize{ \textit{with} $\mathcal{VXP}$}& 
100.0 \% & 
69.1 \% & 
0 &
0 &
0 &
0 &
0 &
1.0 &
1.0 \\\hline

\hline
\textbf{\sffamily \scriptsize The Drive}&
\scriptsize{ \textit{without} $\mathcal{VXP}$}&
59.8 \% &
76.7 \% &
0 &
345 &
18 &
0 &
18 &
1.0 &
1.0 \\

\tiny{(627 frames, 2 targets)} &
\scriptsize{ \textit{with} $\mathcal{VXP}$}& 
79.7 \% & 
76.6 \% & 
0 &
182 &
1 &
0 &
1 &
1.0 &
1.0 \\\hline

\hline
\textbf{\sffamily \scriptsize MOT2016 - Venice-2 }&
\scriptsize{\textit{without} $\mathcal{VXP}$}&
6.4 \% &
69.9 \% &
47 &
27137&
153 &
4 &
150 &
0.987 &
0.241 \\

\tiny{(600 frames, 74 targets)} &
\scriptsize{ \textit{with} $\mathcal{VXP}$}& 
8.1 \% & 
65.2 \% & 
216 &
26535 &
86 &
27 &
77 &
0.946 &
0.486 \\\hline

\end{tabular}

\end{center}
\caption{\small\sffamily \textbf{Evaluation of Tracking Performance}: false positives (FP), misses (M), miss-matches (MM),  non-recoverable miss-matches (non-r. MM), recoverable miss-matches (r. MM), track precision (TP), track recall (TR) 
}
\label{tbl:MOT_result}
\end{table*}%

\subsection{Evaluating Visual Explanations}

We evaluate the generated visual explanations based on their ability to generate low-level object tracks. Towards this we compare the accuracy and precision of the movement tracks the hypothesised event sequences are grounded in.

\smallskip

\subsubsection{Multi-Object Tracking} \quad 
For evaluating the precision and accuracy of the abduced object tracks we follow the ClearMOT evaluation schema for evaluating multi-object tracking performance as described in \citep{Bernardin2008}. 

\begin{itemize}

\item\textit{MOTA}\quad describes the accuracy of the tracking, taking into account the number of missed objects / false negatives (FN), the number of false positives (FP), and the number of miss-matches (MM).

\item\textit{MOTP}\quad describes the precision of the tracking based on the distance of the hypothesised track to the ground truth of the object it is associated to.

\end{itemize}

\noindent These metrics are used to assess how well the generated visual explanations describe the low-level motion in the scene.

\subsection{Results \& Discussion}

We present results of the presented approach for abducing visual explanations ($\mathcal{VXP}$ ) on improving multi-object tracking performance using selected scenes from the  Movie Dataset  and the MOT 2016 Dataset.
Overall the results show that using our proposed method can increase accuracy (\textit{MOTA}) of the tracking. However, the precision (\textit{MOTP}) of the tracking is dropping a little, which is a result of the interpolation, which is not as precise as the detections.

\smallskip

\noindent \textbf{Movie Dataset} \quad 
The scenes in the Movie Dataset contain relatively controlled scenes with few targets. Results on these scenes show that the presented approach can abduce correct event sequences and is capable of correcting many of the errors normally occurring in  multi-object tracking tasks, e.g., fragmented object tracks, id-switches, etc. I.e., the object tracks obtained from the high-level event sequences improve the accuracy (\textit{MOTA})  of the tracking (see Table \ref{tbl:MOT_result}).

\smallskip

\noindent \textbf{MOT Dataset} \quad 
The results for the visual tracking on the \emph{Venice-2} file from the MOT2016 dataset (see Table \ref{tbl:MOT_result}) show, that our approach is capable of dealing with complex data in challenging settings. For comparability we use the DPM based detections provided with the dataset for our evaluation. These detections suffer from a large number of false positives and negatives. Due to this our underling tracking method is only capable of tracking a small part of the overall targets in the data, resulting in a low \textit{MOTA} score. Even so our results demonstrate that high-level abduction can be used to improve tracking performance, i.e., improved \textit{MOTA} by $1.7\%$, and number of miss-matches reduced by $43.8\%$.


%
\smallskip

\noindent Based on the promising results presented in this paper, using basic tracking by detection, we suppose that ASP-based visual explanations can also be used to improve multi-object tracking using more elaborate tracking approaches, e.g., based on continuous energy minimization \citep{Milan16} or minimum cost multi-cuts \citep{Tang17}.

\smallskip

\noindent For the examples presented in this paper, optimal answer-sets are computed rather fast, e.g., for the scene depicted in Figure \ref{fig:qualitative_results}, 457 optimal models are abduced in 0.973s, of which the first model is found after $0.04$s and the last one after $0.93$s.\footnote{We computed hypotheses using a Intel Core i5-4210M 2.60GHz CPU with 12 GB RAM running Ubuntu 16.04}  For longer scenes or in online situations, visual explanations would naturally have to be computed incrementally, as the number of abduced hypothesis grows exponentially with the number of tracks.


\section{Related Work}
Recently, Answer Set Programming (ASP) has become a widely used tool for abductive reasoning and non-monotonic reasoning in general. The work presented in this paper aims at bridging the gap between high-level formalisms for logical abduction and low level visual processing, by tightly integrating qualitative abstractions of space and time with the underlying numerical representations of spatial change. The significance of abducing high-level explanations in a range of contexts has been well established in AI and KR, e.g. in planning and process recognition \citep{Kautz1986,Kautz1991}, vision and abduction \citep{Shanahan05}, probabilistic abduction \citep{Blythe11} etc. Within KR, reasoning about spatio-temporal dynamics on the basis of an integrated theory of space, time, objects, and position \citep{galton2000} or defined continuous change using 4-dimensional regions in space-time has also received significant theoretical  interest \citep{muller1998,Hazarika2002}. \citet{Dubba15} uses abductive reasoning for improving learning of events in an inductive-abductive loop, using inductive logic programming (ILP). The role of visual commonsense in general, and answer set programming in particular, has been used in conjunction with computer vision to formalise general rules for image interpretation in the recent works of \citet{aditya2015visual}. From the viewpoint of computer vision research there has been an interest to synergise with cognitively motivated methods \citep{Aloimonos201542}; in particular the research on {semantic interpretation of visual imagery} is relevant to this paper, e.g., for combining information from video analysis with textual information for understanding events and answering queries about video data \citep{DBLP:journals/ieeemm/TuMLCZ14}, and perceptual grounding and inference \citep{compositional-grounding-jair2015}.

\section{Summary and Outlook}
The paper presents a robust, declarative, and generally usable hybrid architecture for computing visual explanations with video data. With a focus on abductive reasoning in the context of motion tracking, the architecture has been formalised, fully implemented, evaluated with two diverse datasets: firstly, the benchmark MOTChallenge (evaluation focus), and secondly a Movie Dataset (demonstration focus). 


The overall agenda of the work in this paper is driven by a tighter integration of methods in KR and Computer Vision on the one hand, and the twin concepts of  ``\emph{deep semantics}'' \& ``\emph{explainability}'' on the other.  \VXP{} is rooted in state of the art methods in knowledge representation and reasoning (i.e., answer set programming), and computer vision (detection based object tracking, optical flows, RCNN). The overall system is designed to be a part of a larger perception module within autonomous systems, and cognitive interaction systems. The scope of  \VXP{}\ may be further expanded, e.g., for visuo-spatial learning (with \emph{inductive logic programming}), ontological reasoning (with \emph{description logics}), are achievable depending on the scope and complexity of the low-level visual signal processing pipeline, and chosen high-level commonsense knowledge representation and reasoning method(s) at hand.

\small
\nocite{Bhatt2012,DBLP:journals/scc/BhattL08,DBLP:journals/scc/BhattGWH11,Suchan2017-ROBOT}

\bibliographystyle{named}



\newpage


\section{Acknowledgements}{
We acknowledge funding as part of the German Research Foundation (DFG) CRC 1320 ``EASE -- Everyday Activity Science and Engineering'' (http://www.ease-crc.org/) Project P3:  ``Spatial Reasoning in Everyday Activity''. We also acknowledge the Polish National Science Centre project 2016/23/N/HS1/02168.}

\medskip
\medskip
\medskip
\medskip

\section*{{\sffamily\color{blue}\uppercase{Supplementary Material For}}:\\[3pt]\uppercase{}}

{\small\color{black!80!white} J. Suchan., M. Bhatt, Walega, P., Schultz, C. (2018). Visual Explanation by High-Level Abduction: On Answer-Set Programming Driven Reasoning about Moving Objects. In AAAI 2018: Proceedings of the Thirty-Second AAAI Conference on Artificial Intelligence, February 2-7, 2018, New Orleans, USA.}

\medskip
\medskip
\medskip
\medskip
\medskip

\section*{CONTENTS}

\medskip
\medskip

{

We have submitted the following supplementary material with the paper ({\small\color{blue}S1 -- S3}):

\medskip
\medskip

\begin{enumerate}
{	\sffamily\small

	\item [{\small\color{blue}S1}.] A Sample Input Data File
	
	\medskip	
	\item [{\small\color{blue}S2}.] Complete ASP Program

	\medskip
	\item [{\small\color{blue}S3}.] Additional Examples from the Vision MOT Dataset etc (Figs. \ref{fig:dataset_1}-\ref{fig:dataset_3} in the supplementary part)

}
\end{enumerate}

}

\medskip
\medskip
\medskip


\medskip
\medskip

\newpage

\section*{{\sffamily\color{blue}\uppercase{Supplementary 1}}:\\[3pt]\uppercase{Input Data}}

\medskip
\medskip

\textbf{Input Data}\quad This includes movement tracks, consisting of media properties, borders of the frame, and object tracks represented as boxes with $x$, $y$, $width$, and $height$, and the time $t$ given by the frame number with the ASP Program.



\medskip
\medskip

\noindent\includegraphics[width = \columnwidth]{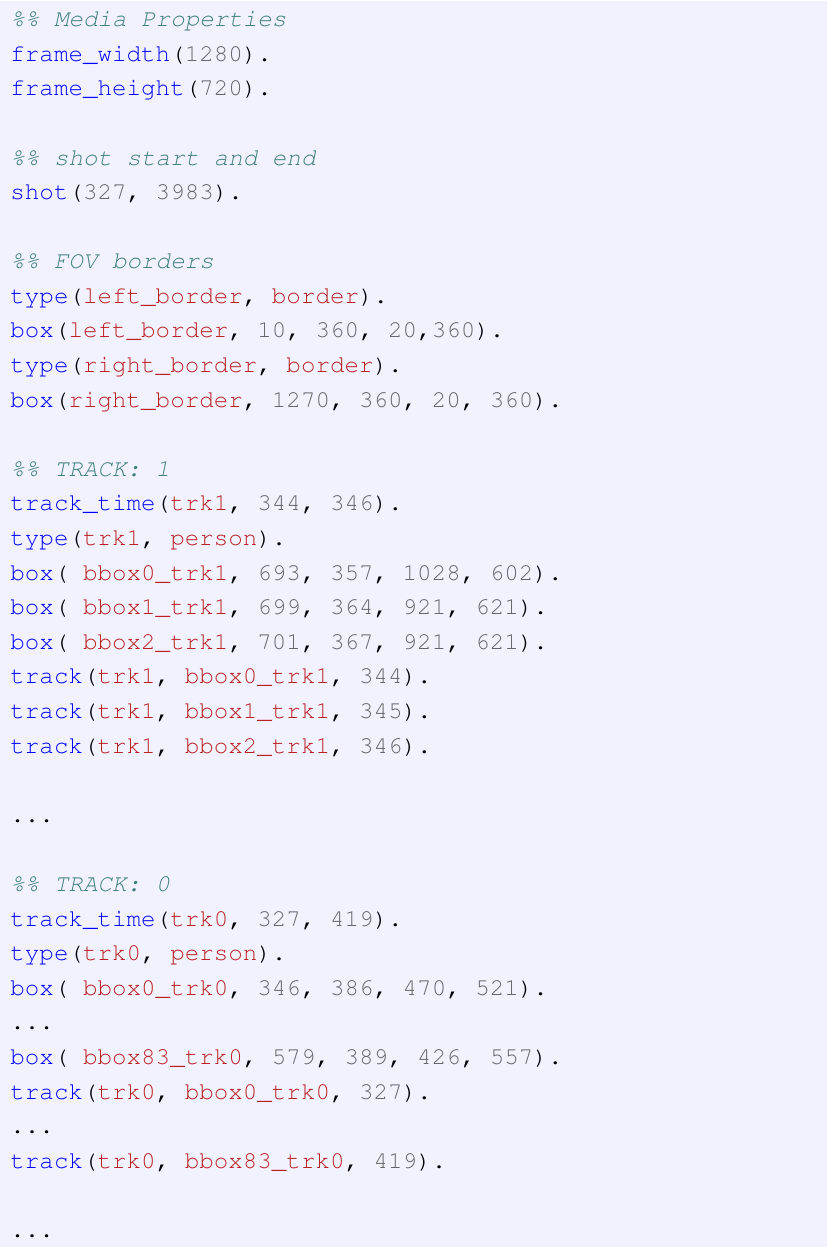}


%
%
%
%
%
%
%
\normalsize


\newpage

\section*{{\sffamily\color{blue}\uppercase{Supplementary 2}}:\\[3pt]\uppercase{Answer Set Programming Based Hypotheses Generation}}


\medskip


\textbf{Abducing Movement Events \& Beliefs}\quad  ASP code for abducing events and beliefs on motion tracks.

\medskip

\noindent Properties of the tracks, i.e., \emph{start}, \emph{end} and \emph{duration}.

\smallskip

\noindent\includegraphics[width = \columnwidth]{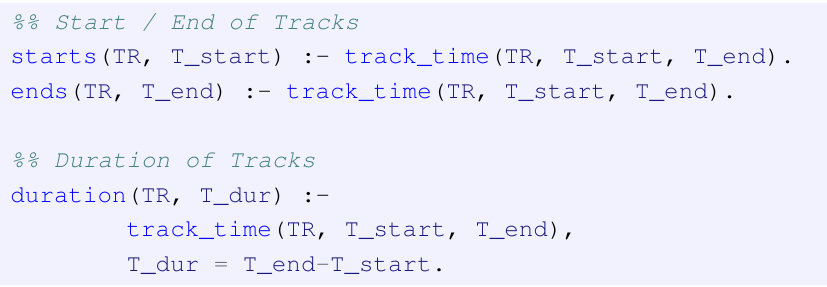}

\smallskip

%
%
%
%

\medskip

\noindent Choice rules for generating explanations on observed changes in the tracks, i.e., \emph{starting} and \emph{ending} of a track.

\smallskip

\noindent\includegraphics[width = \columnwidth]{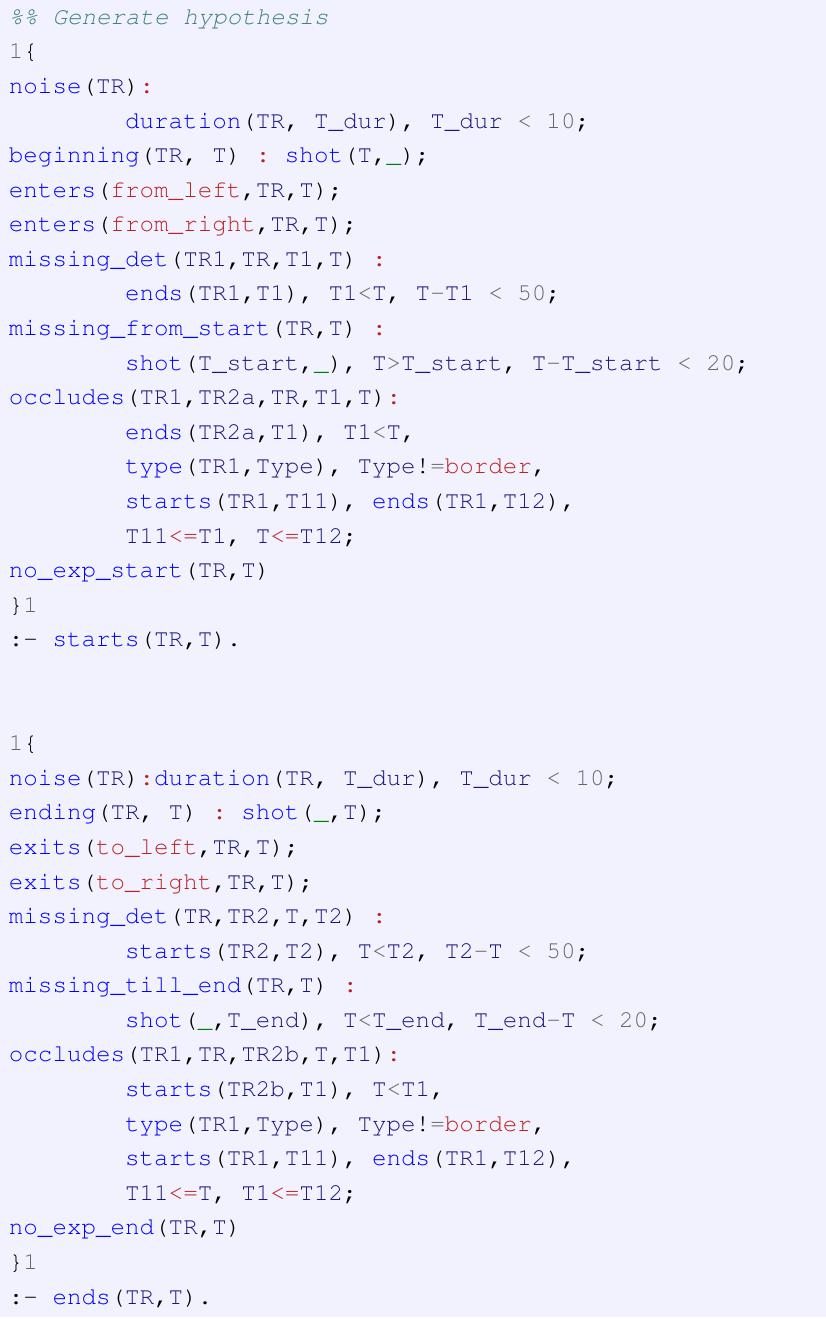}

\smallskip

%
%
%
%

\medskip

\pagebreak

\noindent Spatial preconditions for abducable events using topological relations from the region connection calculus.

\smallskip

\noindent\includegraphics[width = \columnwidth]{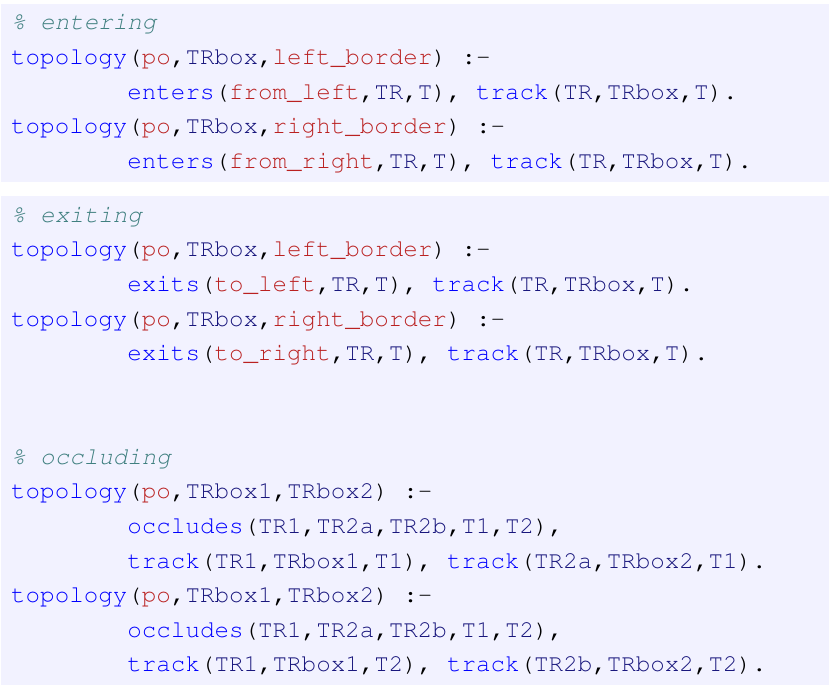}

\smallskip

%
%
%
%
%


\medskip
\medskip

\noindent\textbf{Multi-Objective Optimization}\quad ASP based optimization of generated hypotheses using cost functions for abduced events and object tracks.

\medskip

\noindent Weights for events and motion.

\smallskip

\noindent\includegraphics[width = \columnwidth]{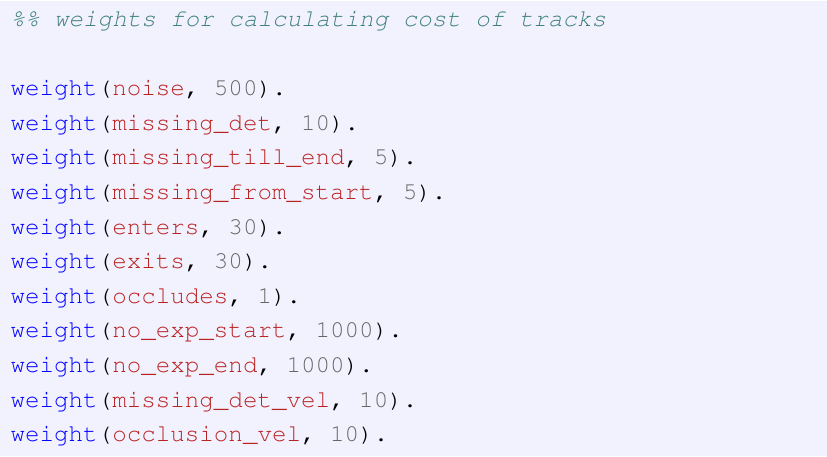}

\smallskip

%
%
%
%
%
%
%

\medskip

\noindent Costs for events based on event type and length.

\smallskip

\noindent\includegraphics[width = \columnwidth]{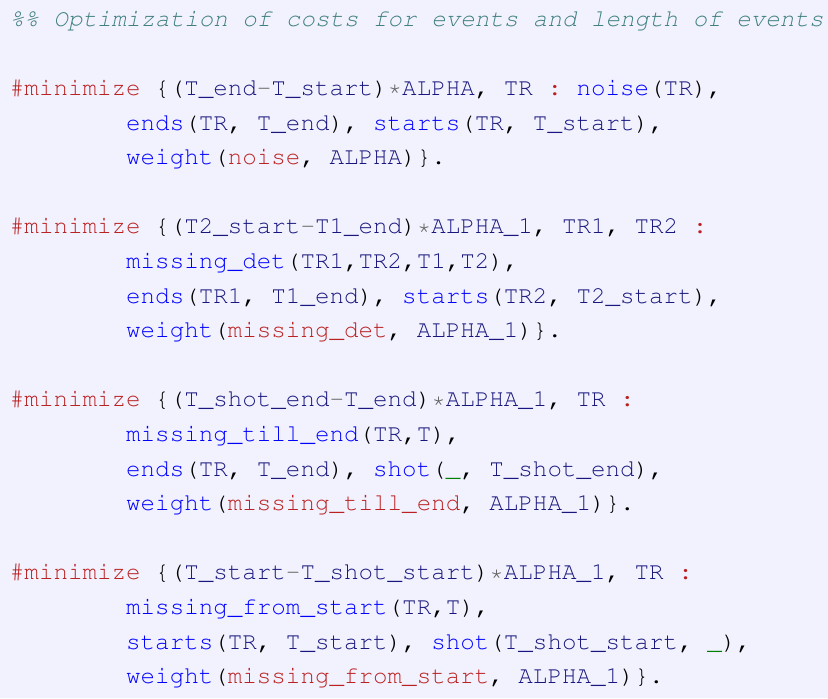}

\smallskip

\newpage

\smallskip

\noindent\includegraphics[width = \columnwidth]{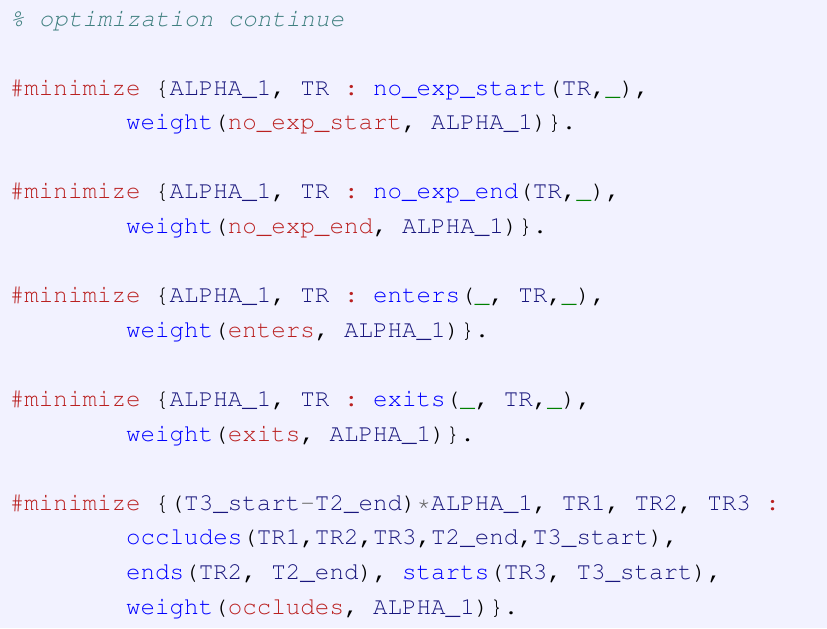}

\smallskip

%
%
%
%
%
%
%
%
%
%
%
%
%
%

\medskip

\noindent Cost for abduced motion tracks.

\smallskip

\noindent\includegraphics[width = \columnwidth]{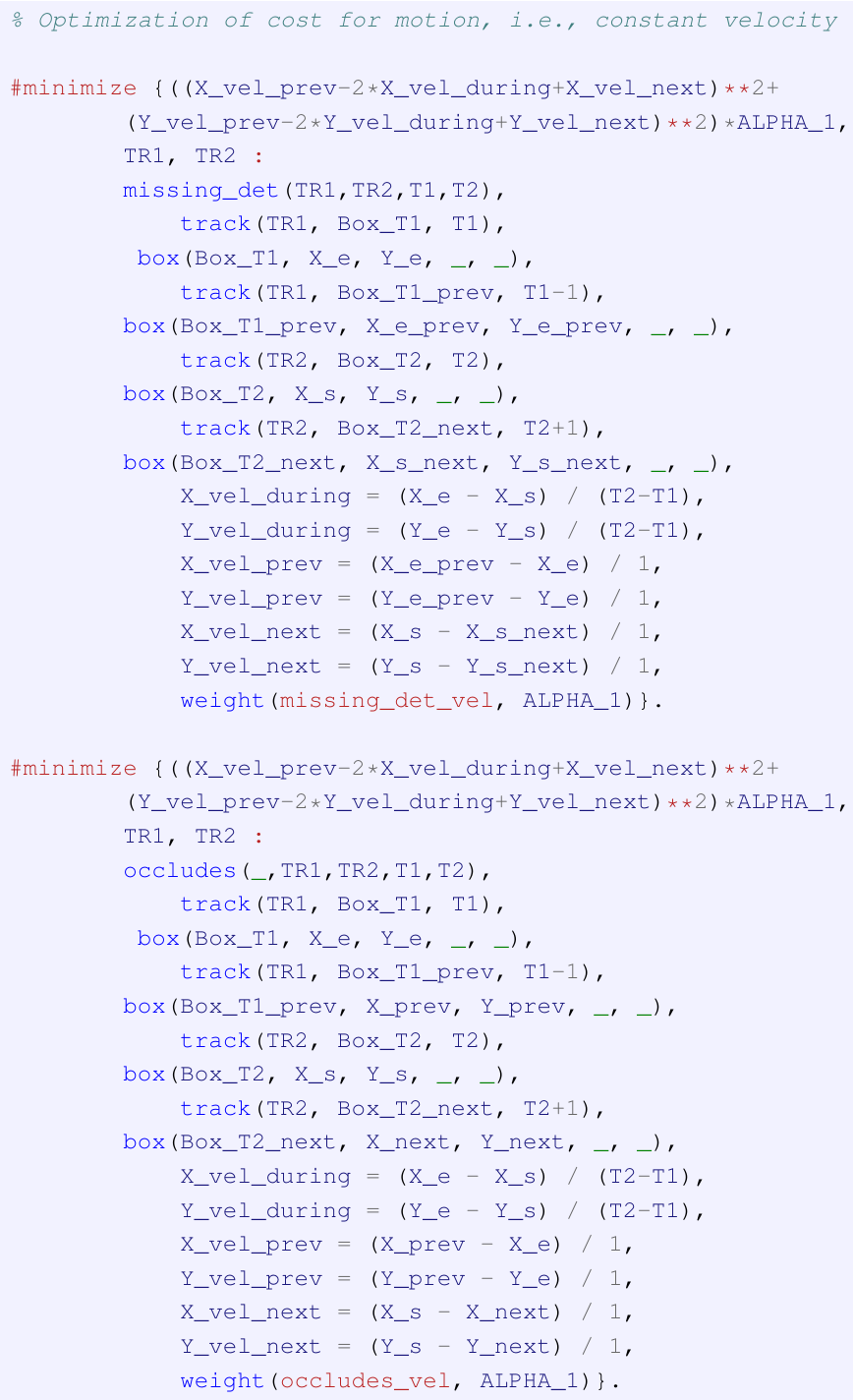}

\smallskip

%
%
%

\pagebreak

\section*{{\sffamily\color{blue}\uppercase{Supplementary 3}}:\\[3pt]\uppercase{Sample Scenes Illustrating VXP}}



\begin{figure*}[t]
 
\centering
\includegraphics[width = 0.95\textwidth]{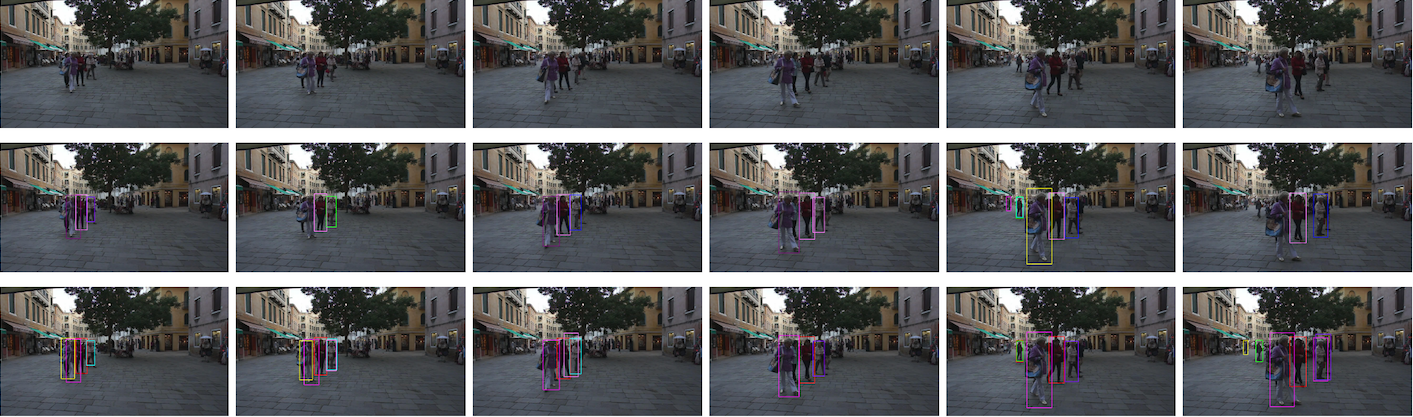}

\caption{{Sample Scene from the MOT dataset: corrected tracks in challenging setting (noisy detections and many targets). First Row: original video. Second Row: Object Tracks obtained from Tracking. Third Row: Corrected People Tracks.}}
\label{fig:dataset_1}
\end{figure*}

\begin{figure*}[t]
 
\centering

\includegraphics[width = 0.95\textwidth]{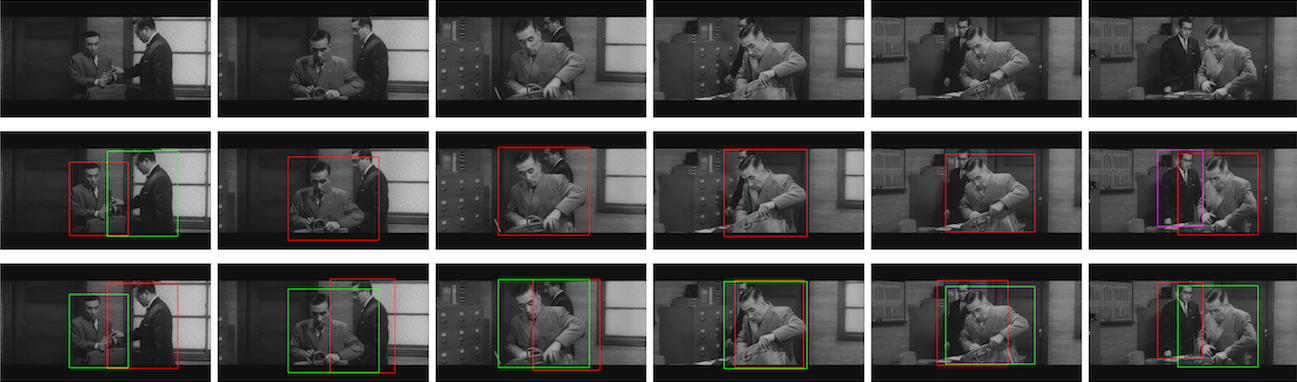}

\caption{{Sample Scene from the movie ``The Bad Sleep Well'' by Akira Kurusawa (1960): abduced high-level event passing behind and corrected people tracks. First Row: original video. Second Row: Object Tracks obtained from Tracking. Third Row: Corrected People Tracks.}}
\label{fig:dataset_2}
\end{figure*}

\begin{figure*}[t]
 
\centering

\includegraphics[width = 0.95\textwidth]{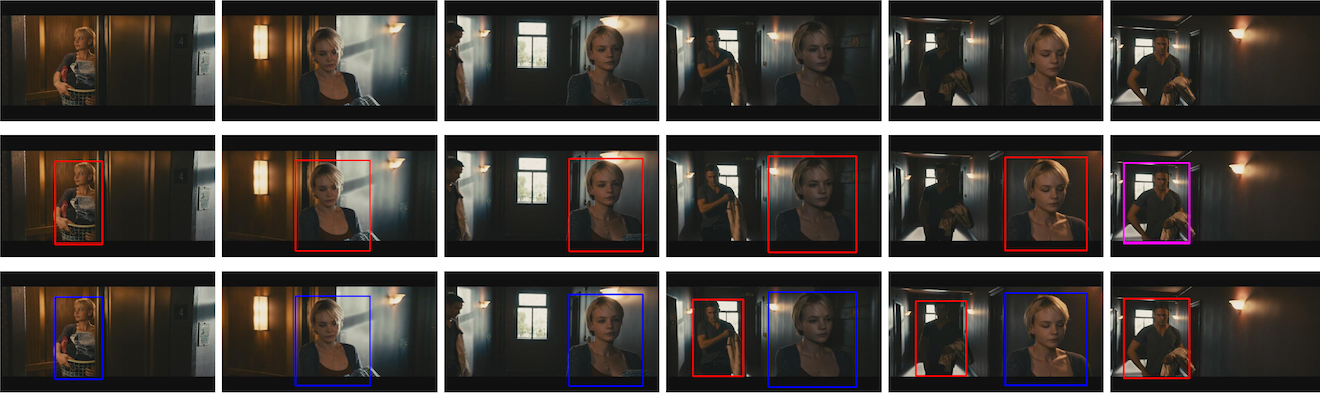}

\caption{{Sample Scene from the movie ``Drive'' by Nicolas Winding Refn  (2011): abduced missing detections and corrected people tracks. First Row: original video. Second Row: Object Tracks obtained from Tracking. Third Row: Corrected People Tracks.}}

\label{fig:dataset_3}
\end{figure*}

Additional Figures \ref{fig:dataset_1}-\ref{fig:dataset_3} (in this supplementary part) provide sample results for corrected people tracks based on abduced visual explanations from the movie dataset and the MOT dataset.


\medskip
\medskip

%
%
%
%
%
%
%
%

\end{document}